\documentclass[10pt, letterpaper]{article}
\usepackage{graphicx}
\usepackage[super]{nth}
\usepackage{soul}
\usepackage[dvipsnames]{xcolor}

\usepackage{cogsci}

\cogscifinalcopy 
\usepackage{pslatex}
\usepackage{apacite}
\usepackage{float} 
   
\usepackage{tipa}
\usepackage{enumitem}
\setlist{nosep}

\title{A Neural Network Model of Lexical Competition \\during Infant Spoken Word Recognition}
 
\author{{\large \bf Mihaela Duta (mihaela.duta@psy.ox.ac.uk)} \\
  University of Oxford, Department of Psychology\\ Anna Watts Building, Radcliffe Observatory Quarter, Woodstock Road \\ Oxford OX2 6GG, United Kingdom
  \AND {\large \bf Kim Plunkett (kim.plunkett@psy.ox.ac.uk)} \\
  University of Oxford, Department of Psychology,\\ Anna Watts Building, Radcliffe Observatory Quarter, Woodstock Road \\ Oxford OX2 6GG, United Kingdom}

\begin{document}

\maketitle

\begin{abstract}

Visual world studies show that upon hearing a word in a target-absent visual context containing related and unrelated items, toddlers and adults briefly direct their gaze towards phonologically related items, before shifting towards semantically and visually related ones. We present a neural network model that processes dynamic unfolding phonological representations and maps them to static internal semantic and visual representations. The model, trained on representations derived from real corpora, simulates this early phonological over semantic/visual preference. Our results support the hypothesis that incremental unfolding of a spoken word is in itself sufficient to account for the transient preference for phonological competitors over both unrelated and semantically and visually related ones. Phonological representations mapped dynamically in a \emph{bottom-up} fashion to semantic-visual representations capture the early phonological preference effects reported in a visual world task. The semantic-visual preference observed later in such a trial does not require \emph{top-down} feedback from a semantic or visual system.

\textbf{Keywords:} 
language; neuro-computational models; development; visual world task; phonology; semantics; cohort effects; machine learning; lexical competition; spoken word recognition; attention.
\end{abstract}

\section{Introduction}

Upon hearing a spoken word, listeners selectively attend to an item that best matches the word's referent. For example, on seeing a display containing a hat and a bear, listeners selectively attend to the hat when they hear \emph{trousers}. Likewise, they selectively attend to a picture of a train upon hearing \emph{trousers} when presented with a train and a fridge. 

\begin{figure}[h]
\begin{center}
\begin{tabular}{cc}
\includegraphics[width=0.225\textwidth]{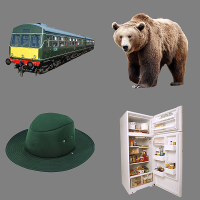} &
\includegraphics[width=0.225\textwidth]{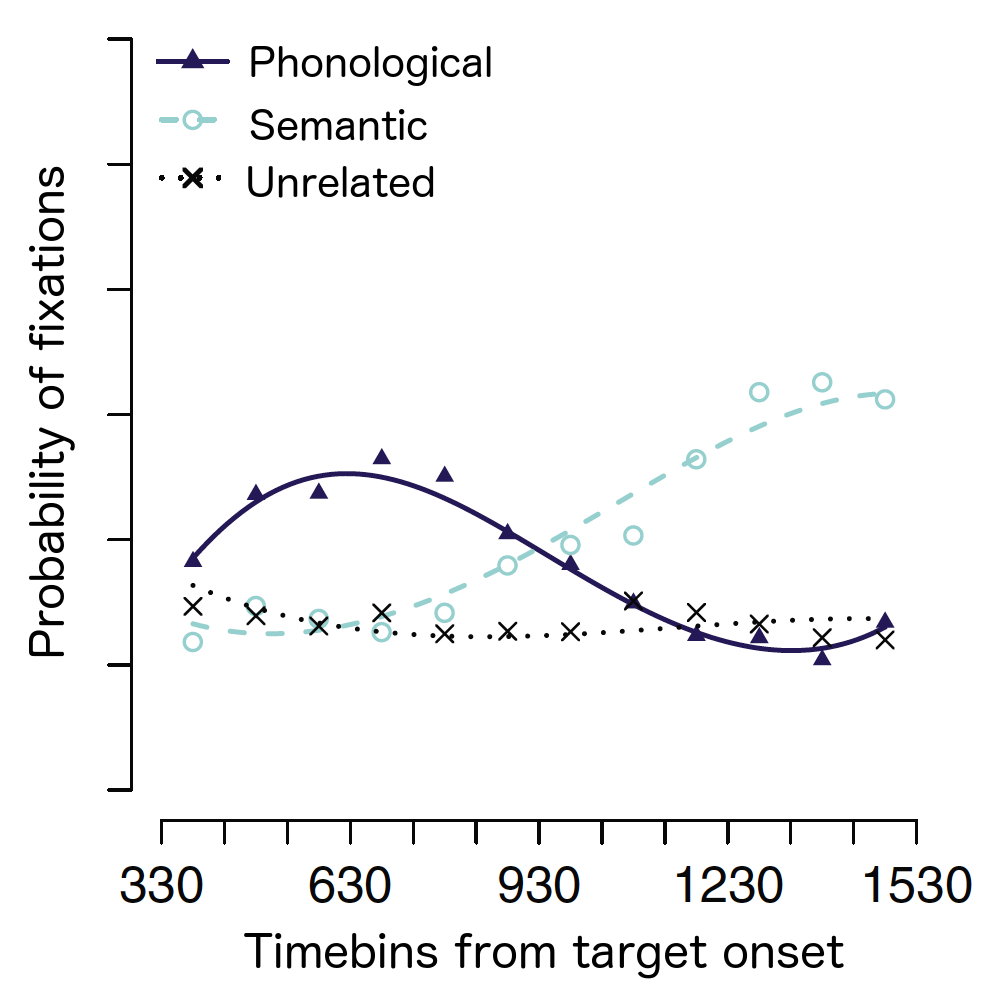} \\
(a) & (b)\\
\end{tabular}
\end{center}
\caption{ (a) Example of the type of display used in visual world tasks \cite{huettig2007tug, chow2017spoken} and (b) Successive fixation of phonological and semantic foils in a 4-picture visual world task by 30-month old toddlers \cite{chow2017spoken}.}
\label{fig:VWP}
\end{figure}

In more complex displays such as Figure~\ref{fig:VWP}(a), which contain both phonological and semantic foils to the referent of \emph{trousers}, listeners exhibit selective attention to both types of foil relative to the unrelated items. Furthermore, listeners selectively and briefly attend to the phonological foil \emph{before} switching attention to the semantically related item. Figure~\ref{fig:VWP} (b) depicts early fixations to phonological foils by 30-month old toddlers within 400ms of word onset followed by a shift to semantic foils \cite{chow2017spoken}. Similar results are found with adults, though the initial phonological preference is conditioned by the picture preview time relative to word onset \cite{huettig2007tug}.

This pattern of findings is explained by assuming that the listener generates a phonological representation from the unfolding auditory signal and uses this representation to identify the best matching semantic and visual representation generated from the visual input provided by the images. The locus of the match could, in principle, occur at any of the representational levels linking the auditory and visual stimuli: phonological, semantic or visual. However, the early preference for the phonological foil suggests that the locus of the match resides at the phonological level\footnote{\citeA{huettig2007tug} also point out that removal of the picture preview phase in this task obliterates the early phonological preference, presumably because participants don't have time to generate the phonological codes for the images.\label{huettig}}.

A recent computational model uses a hub-and-spoke architecture to capture the integration of phonological, semantic and visual information in driving attention in visual world tasks \cite{smith2017multimodal}. The recurrent hub of the model 
receives inputs from visual and phonological layers, and propagates activity to target semantic and eye layers which themselves feedback activity to the hub. Using an artificially constructed corpus, the model successfully replicates rhyme effects, e.g., hear \emph{coat} and look at boat \cite{allopenna1998tracking}. 

\citeA{smith2017multimodal} argue that the close integration of visual, phonological and semantic information in the hub is central to the model's capacity to capture the phonological rhyme effect observed in visual world tasks. We would argue that a feature of the model also critical for obtaining a preference for rhyming over unrelated items is the persistence of all the discrete phonological segments at the input during processing. The rhyming segment of the word thereby comes to dominate the phonological input as the simulation of a visual world trial proceeds. 

In this paper, we explore the hypothesis that incremental unfolding of the spoken word, one phonological segment at a time, is sufficient in itself to account for early phonological preferences of the type depicted in Figure~\ref{fig:VWP}(b), i.e., a transitory early preference for phonologically related items over \emph{both} semantically and visually related items, as well as unrelated ones, followed by a preference for semantically and visually related items over \emph{both} unrelated and phonologically related ones. We evaluate this hypothesis by constructing a neural network model that  processes \emph{only} unfolding phonological representations of words at the input and learns to map these dynamic phonological sequences to corresponding static semantic and visual representations of the words' referents at the output. In essence, the model can be considered to implement a form of lexical comprehension. Particularly noteworthy aspects of the model include:
\begin{itemize}
    \item All representations used in the model are `naturalistic' insofar as they have been derived from real corpora.
    \item The model's vocabulary is derived from a realistic toddler vocabulary taken from parental questionnaire studies.
    \item The phonological input consists of dynamic, as opposed to static slotted representations. The model itself builds embedded representations of the unfolding word using gated recurrent units (GRUs).
\end{itemize}{}
As a first step, we focus on phonological \emph{onset} effects with a view to extending the model eventually to encompass phonological \emph{rhyme} effects, \protect\`{a} la \protect\citeA{smith2017multimodal}. To anticipate the findings, our model successfully accommodates the early phonological over semantic/visual preference observed in visual world studies \cite{huettig2007tug, chow2017spoken}. However, we do \emph{not} consider this model a complete account of language mediated attention in visual world settings, but rather a tool to explore the power of dynamic phonological representations in guiding our attention to semantic and visual items.

\section{Methods}

The software was developed in \emph{Python 3} using \emph{numpy}, \emph{scipy} and \emph{pandas} libraries and models were implemented, trained and simulated with the \emph{pytorch} machine learning framework \cite{pytorch}. 

\subsection{Vocabulary}
\begin{figure}[h]
\begin{center}
\begin{tabular}{l}
\includegraphics[width=0.45\textwidth]{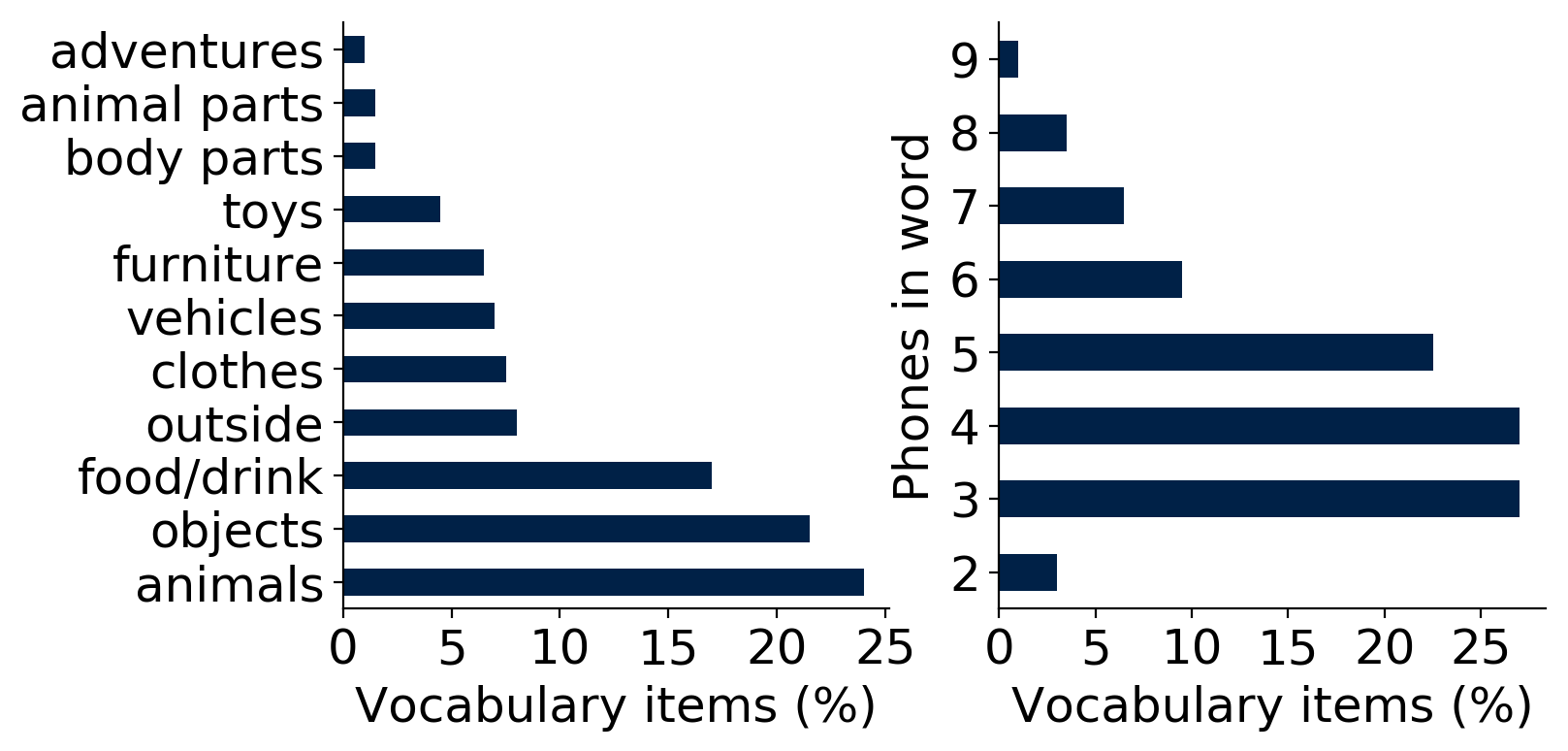}\\
\includegraphics[width=0.45\textwidth]{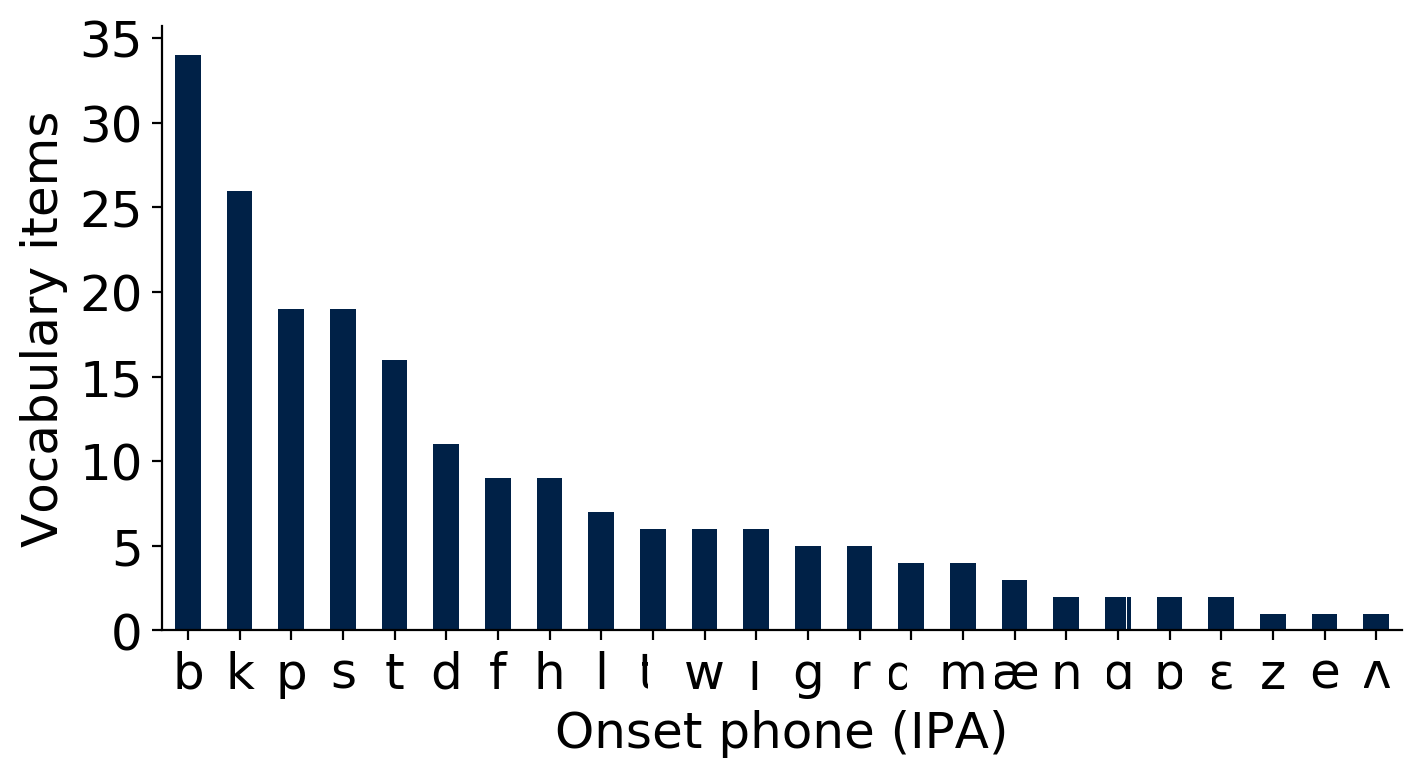}
\end{tabular}
\end{center}
\caption{Descriptive statistics for vocabulary items: item distribution across semantic categories, word length distribution and cohort size distribution across phones.}
\label{fig:corpus_stats}
\end{figure}

The corpus consists of 200 imageable noun items from the infant lexicon, as documented by the Oxford Communicative Development Inventory data \cite{hamilton2000infant}. Vocabulary items come from 11 distinct semantic categories, with a majority (62\%) belonging to the categories of animals, food/drink or household objects. Labels range in length from 2-phone to 9-phone words, 94\% of which start with a  consonant and 6\% with a vowel. The phone inventory of the corpus consists of 39 distinct phones, 26 consonants and 13 vowels. Of the 189 items with a consonant onset label, 66\% have a cohort larger than 5 items and start with \emph{b}, \emph{k}, \emph{p}, \emph{s}, \emph{t} or \emph{d} phones. Figure \ref{fig:corpus_stats} gives distribution plots for category membership, label length and onset phone identity across the entire corpus. 

\subsection{Phonological representations}
Each phone in the inventory is assigned a feature-based distributed binary encoding based on 20 articulatory and phonological features  \cite{karaminis2018effects}. The phonological representation for each vocabulary item is then constructed as the sequence of feature representations of its phones in the order in which they appear as the spoken word unfolds. Eight items in the corpus have labels embedded in at least one other longer vocabulary item (see Table~\ref{tab:embed}). A segmentation character for which all 20 phonological features are set to 1 was introduced to mark the offset of all labels. 
To account for phone co-articulation, the transition between consecutive phone representations is achieved via two intermediate vectors so that the transition between the feature values $1$ to $0$  consists of two intermediate values of $0.95$ and $0.05$ and \emph{vice versa}. 

\begin{table}[h]
\begin{center} 
\caption{Items with labels embedded in other items' labels.} 
\label{tab:embed} 
\vskip 0.12in
\begin{tabular}{ll} 
\hline
Embedded label & Embedding labels \\
\hline\hline
bee [b\textipa{i:}] & beach [b\textipa{i:}\textteshlig], beans [bi:nz]\\
doll [d\textipa{6}l] & dolphin [d\textipa{6}lf\textipa{I}n]\\
glass [gl\textipa{A:}s] & glasses [gl\textipa{A:}s\textipa{@}z]\\
key [k\textipa{i:}] & keyboard [k\textipa{i:}b\textipa{O:}d]\\
cat [k\textipa{ae}t] & caterpillar [k\textipa{ae}t\textipa{@}p\textipa{I}l\textipa{@}]\\
lamb [l\textipa{ae}m] & lamp [l\textipa{ae}mp]\\
tie [t\textipa{A}\textipa{I}] & tiger [t\textipa{A}\textipa{I}g\textipa{@}]\\
tooth [t\textipa{u:}\textipa{T}] & toothbrush [t\textipa{u:}\textipa{T}br\textipa{2}\textipa{S}]\\
\hline
\end{tabular} 
\end{center} 
\end{table}

\subsection{Visual and semantic representations}
The visual representation for each vocabulary item is derived from the response to an illustration of the item of a \emph{resnet18} deep neural network pre-trained on ImageNet,
using the 512-dimensional activation vector for the \emph{avgpool} layer
\cite{he2016deep, pytorch, imagenet_cvpr09}. The semantic representations are 100-dimensional word vectors from the GloVe model pre-trained on aggregated global word-word co-occurrence statistics from a 6 billion token corpus composed of the Gigaword5 and Wikipedia 2014 dump \cite{pennington2014glove}.

\begin{figure}[h]
\begin{center}
\includegraphics[width=0.4\textwidth]{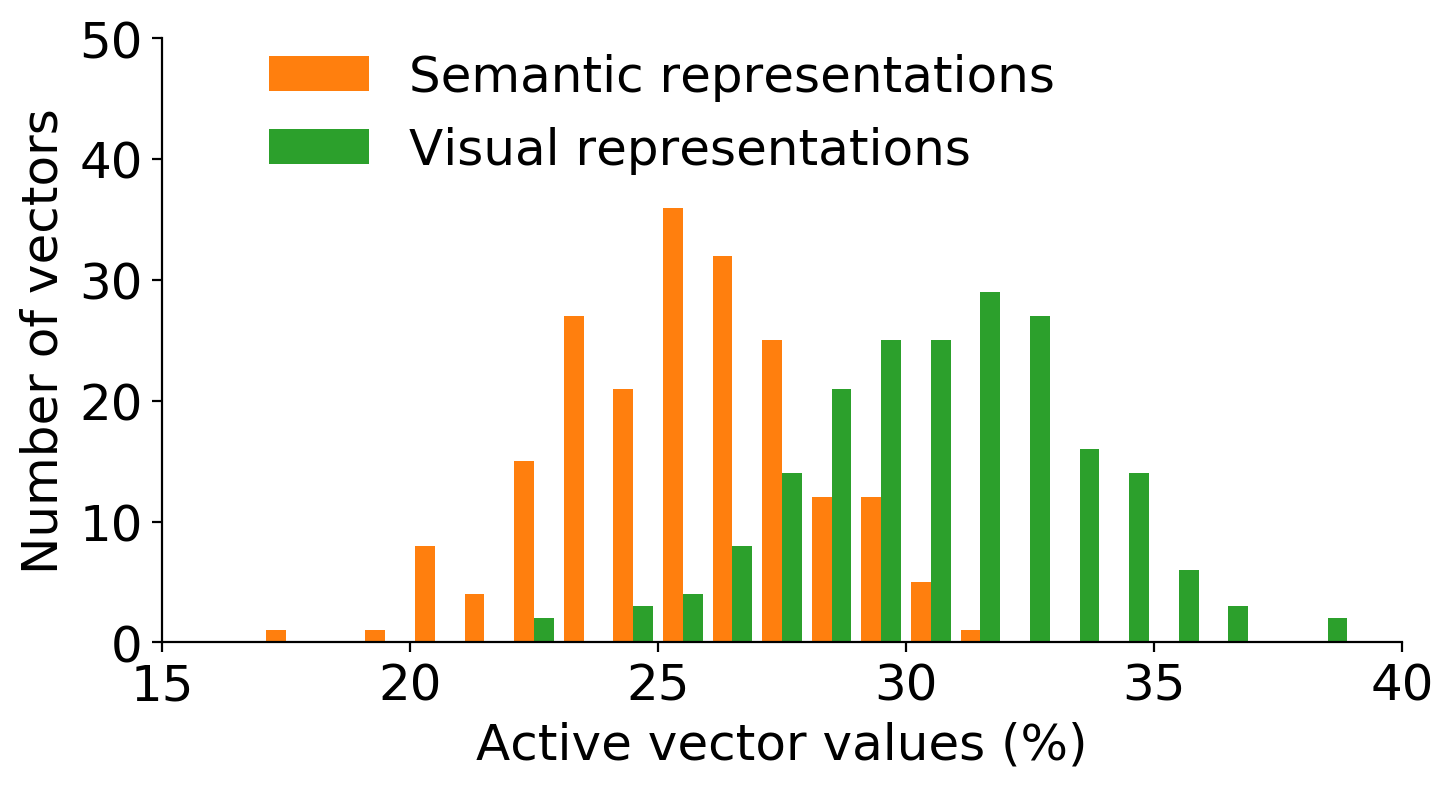} (a)
\includegraphics[width=0.4\textwidth]{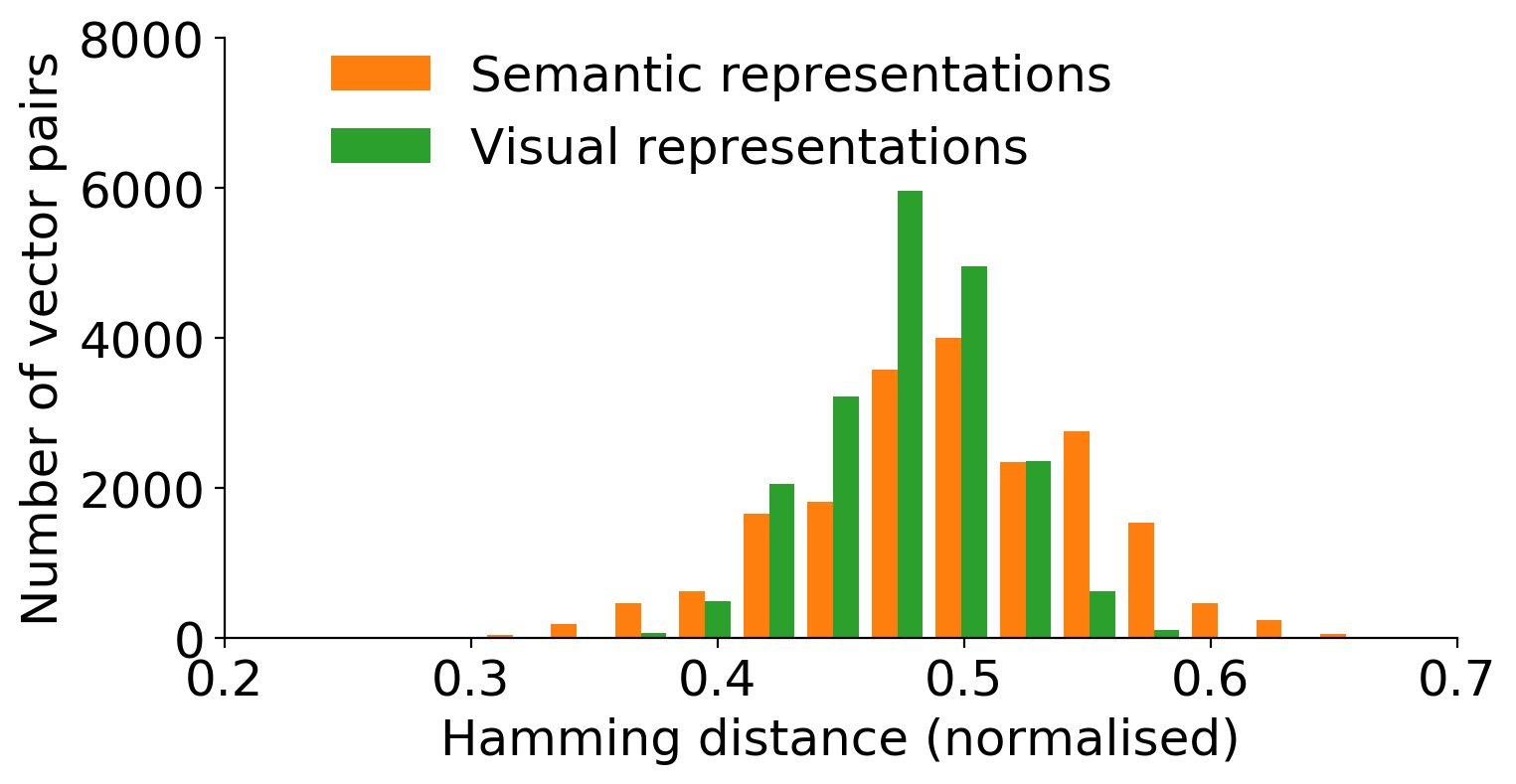} (b)
\end{center}
\caption{(a) Distributions for the active representation vector values (equal to 1) in the semantic and visual representations. (b) Distributions for Hamming distance between pairs of semantic and visual representation vector.}
\label{fig:reps_distr}
\end{figure}

The visual and semantic representation vectors are processed to replace outliers (vector values with zscore $>$ 2) with the median value for the corresponding dimension. 
Visual representation vectors are further processed using principal component analysis to reduce their dimensionality to 150 (cumulative variance explained: 95\%).
Both visual and semantic representation vectors are then rounded to obtain binary vectors, and concatenated to obtain aggregated semantic-visual representation of the items. The distribution plot for the number of active representation values (equal to 1) given in Figure \ref{fig:reps_distr}(a) shows that both semantic and visual representations are sparsely distributed, semantic representations being slightly sparser than visual ones. 

\subsection{Architecture and training}
\begin{figure}[h]
\begin{center}
\includegraphics[width=0.45\textwidth]{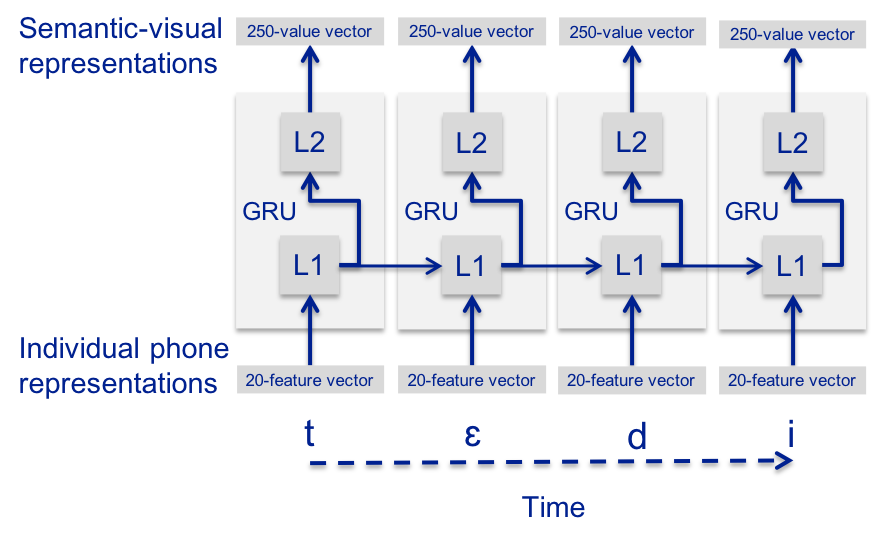}
\end{center}
\caption{Illustration of model activation for the unfolding of the word \emph{teddy}; L1 and L2 are the 1$^{st}$ and 2$^{nd}$ GRU layers; intermediate co-articulation timesteps are suppressed in the graphic.}
\label{fig:model-diagram}
\end{figure}

The model is designed to associate the dynamic unfolding of the phonological representations of the vocabulary items with the corresponding aggregated static semantic and visual representations. To achieve this, the architecture consists of a two-layer gated recurrent unit (GRU) network \cite{cho2014learning} whose inputs and outputs are a 20-dimensional phone encoding vector and a 250-dimensional vector of aggregated semantic-visual representations, respectively (see Figure~\ref{fig:model-diagram}). The processing cycle for an individual vocabulary item consists of the number of timesteps required to fully unfold the phones in the item's label including the intermediate steps accounting for phone co-articulation. 

Training was performed on the entire corpus using batch update and stochastic gradient descent (learning rate: 0.4, momentum: 0.4 and Nesterov momentum enabled \cite{sutskever2013importance}). A training trial consisted of the unfolding at the input of the complete phonological representation of the label of a vocabulary item matched with the corresponding aggregated semantic-visual representations as targets. All training trials had the same number of timesteps required to completely unfold the longest label in the vocabulary. For shorter labels the inputs were padded with zeros from the label offset to the end of the trial. The target semantic-visual representations were active only during label unfolding and were set to zero from label offset to the end of the trial. 

The number of training epochs was set to the one that enabled all trained models to learn all vocabulary items. To evaluate whether a word has been learned, the entire sequence of phones in its label is presented at the input and the normalised Hamming distances from the model output at label offset to the aggregated semantic-visual representations for all the vocabulary items are evaluated. A model is considered to have learned a word if the shortest such distance is to the word's target aggregated semantic-visual representation. 

\subsection{Simulating visual world trials}

The trained models were evaluated in simulations of `target absent' visual world trials in which the output activations of the model are evaluated for referents in a visual display with four candidates: a phonologically-related referent, a semantically-related referent, a visually-related referent and an unrelated referent. At each time step during the unfolding of the target label, the model activation for a candidate referent is calculated as one minus the normalised Hamming distance between the current model output and the referent's aggregated semantic-visual representation. The model is assumed to direct attention to the candidate referent with the highest activation, i.e. the candidate referent whose aggregated semantic-visual representation is closest to the current model output. 

Simulation trials each consisting of an input target label, and a set of phonologically, semantically, visually and unrelated items were constructed as follows:
\begin{itemize}
    \item phonologically related item (PREL): shares the onset phone with the target label, but is both semantically and visually unrelated to the target
    \item semantically related item (SREL):  is semantically related, but visually and phonologically (both onset and rhyme) unrelated to the target
    \item visually related item (VREL): is visually related, but semantically and phonologically (both onset and rhyme) unrelated to the target 
    \item unrelated item (UREL):  is phonologically (both onset and rhyme), semantically and visually unrelated to the target
\end{itemize}
To avoid any accidental bias, any item appeared only once in any of the phonologically related, semantically related, visually related or unrelated referent category. 
Also, to avoid spurious relationships,  no items whose labels were embedded in or embedded other labels were included. 

The selection of the related and unrelated items was made taking into account the normalised Hamming distances between the target and competitor items in the aggregated semantic-visual representation space. An item was considered semantically related or unrelated to the target if the normalised Hamming distance between its semantic representation vector and that of the target was in the top \nth{10} or bottom \nth{25} percentile, respectively. Similarly, for the visually related and unrelated items their distance was in the top 0.5\textsuperscript{th} or bottom \nth{25} percentile, respectively. Figure \ref{fig:reps_distr}(b) shows that the pairwise distances have a wider distribution for the semantic compared to visual representations. Therefore, a stricter top percentile threshold was used for the visual representations to ensure that the corresponding distance threshold was similar across the two representations. A total of 18 trials complying with all these selection criteria could be assembled from the entire corpus. Of the 18 simulation target words, 4 are 3-phones long and the rest are at least 4-phone long.

\section{Results}

\subsection{Model training}

\begin{figure}[h]
\begin{center}
\includegraphics[width=0.34\textwidth]{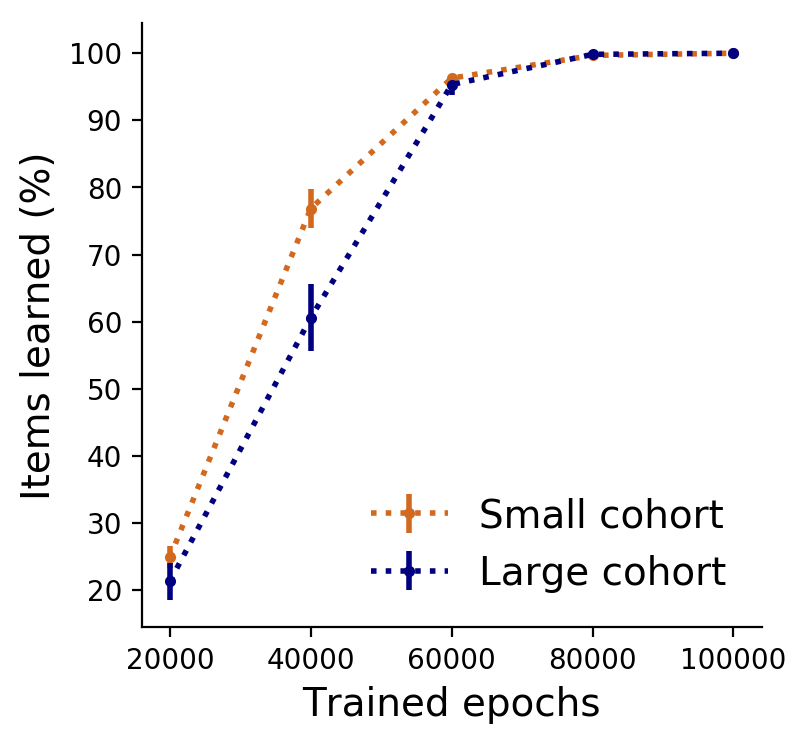}
\end{center}
\caption{Vocabulary size during training: large cohort contain 25 items or more, see Figure \ref{fig:corpus_stats}. Bars: one standard deviation.}
\label{fig:training_vocabulary}
\end{figure}

Twenty models were each trained for 100000 epochs. This allowed all models to learn all the vocabulary items. Vocabulary size for each model was evaluated every 20000 epochs during training. Figure \ref{fig:training_vocabulary} shows that models are faster at learning words with small cohorts, though successful learning of the entire vocabulary is achieved in both cases.

\subsection{Simulations}

\begin{figure}[htbp]
\begin{center}
\includegraphics[width=0.45\textwidth]{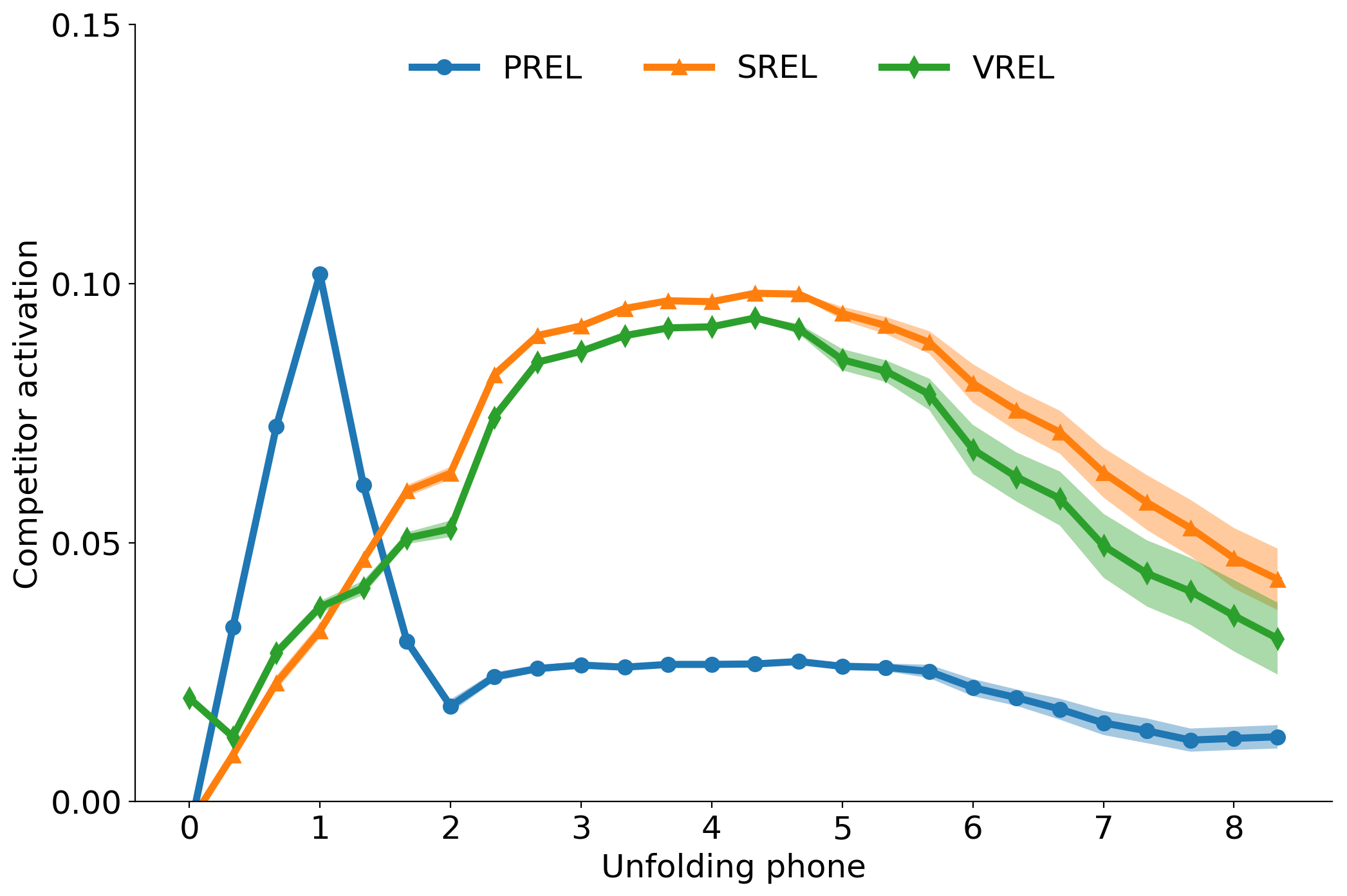}
\end{center}
\caption{ 
Simulation output:  activation time courses as the word unfolds for phonological competitor (PREL), semantic competitor (SREL) and visual competitor (VREL) (ribbons: standard error of the mean).}
\label{fig:vwppsvu_timecourse}
\end{figure}

Figure \ref{fig:vwppsvu_timecourse} plots the outcome of simulating the trained models: the horizontal axis is the simulation timestep from the onset of the target label and the vertical axis is the grand mean of activations for the phonologically-related, semantically-related and visually-related competitors relative to the activation of the unrelated competitor.

Results show that activations for the phonologically related items are larger than any other activation earlier on in the unfolding of the label, shifting to larger activations for semantically and visually related items later in the trial.

\setlength{\tabcolsep}{0pt}
 
\section{Discussion}

The research reported in this paper evaluates the proposal that incremental unfolding of a spoken word is in itself sufficient to account for the transient preference for phonological competitors over both unrelated and semantically/visually related ones in a visual world task. We evaluate this proposal with a neural network model designed to map  dynamic phonological inputs to static semantic-visual representations via gated recurrent units (see Figure~\ref{fig:model-diagram}). 

The 20 trained models each successfully learned the entire set of 200 vocabulary items. The trained models were tested in simulated `target-absent' visual world trials in which the model activations for the four competitor referents --- either unrelated to the referent of the unfolding word, or phonologically, semantically or visually related to it ---are continuously estimated. The activation is estimated by the distance between the current model output and the semantic-visual representations of all the candidate referents.

Figure~\ref{fig:vwppsvu_timecourse} depicts a clear early higher activation of the phonological competitor followed by a shift in favour of the semantic and visual competitors later in the trial. We interpret these activations as an early preference for the phonological competitor in a `target-absent' visual world trial, followed by a later preference for the semantic and visual competitors. These results confirm our proposal that a dynamic unfolding phonological input is sufficient to generate an initial preference for the phonological competitor over \emph{both} semantic and visual competitors in a visual world task.

The models also have the desirable quality of exhibiting a rapid increase in vocabulary during the earlier stages of training, a phenomenon often reported in the child language literature as vocabulary spurt \cite{Mcshane79, mcmurray2007defusing,Plunkett92}. The timing of the spurt is conditioned by the cohort size of vocabulary items. Although we are unaware of any studies specifically investigating the relation between  vocabulary growth and word cohort size, some studies of early lexical development report a deleterious effect of similar sounding words on vocabulary development and lexical processing \cite{swingley2007lexical, mani2011phonological}.

We now turn to the issue of why our model exhibits an early phonological preference over a semantic-visual preference. Upon `hearing' the onset phone of a word, the model output migrates to the region of the semantic-visual space  consistent with the current phonological input. In a `target-absent' visual world trial this is bound to be towards the representation of the phonological competitor---if one is present---which is the only one consistent with the onset phone. Therefore, the phonological competitor has the highest activation. However, as the input word unfolds over time, the semantic-visual region consistent with the phonological input shifts. The model has been trained to associate \emph{words unfolding towards complete forms} with corresponding semantic-visual representations: the more of the word the model `hears', the more its semantic-visual outputs shift towards the semantic-visual associates of the input word. Hence, the models favours phonological competitors before semantic-visual competitors in a `target-absent' visual world task. The model therefore predicts that in such a task where the scene also contains a phonological onset competitor, unambiguous identification of the target would be delayed relative to a scene that did not contain such a competitor. Evidence for such a delay has been reported in infant word recognition experiments. When 24-month-olds were presented with a display containing a phonological onset competitor (doll-dog), their target responses were delayed but not when the pictures’ labels rhymed (doll-ball) \cite{Swingley99}.

It is worth noting that our model architecture does not permit feedback of activity from the semantic-visual representations to the phonological representations. In other words, there is no `implicit naming' of the stimuli in the visual world trial simulations reported: the model does not generate phonological representations from semantic-visual representations. A corollary of this feature is that the locus of the match between auditory and visual stimuli in a visual world task lies at the semantic-visual level, not at the phonological level. This built-in assumption of the model is at odds with the claim that reducing picture preview time in a visual world task can eliminate early phonological preferences (see \citeNP{huettig2007tug}). However, we note a growing body of empirical evidence that an extended picture preview time is not required to observe an early phonological preference effect in visual world tasks \cite{villameriel2019language, rigler2015slow}. These recent findings point to the possibility that other task demands that highlight semantic competitors may suppress phonological effects during referent identification.

Some forms of semantic feedback, such as that implemented in \citeA{smith2017multimodal}, may serve to eliminate  early phonological preferences in visual world tasks in certain circumstances, such as those reported by \citeA{huettig2007tug}. In this case, identification of the neuro-computational mechanism(s) responsible for controlling the presence/absence of the widely-reported phonological effects would be required. We speculate that growth in \emph{top-down} connectivity from semantic representations, perhaps through the emergence and consolidation of the lexical-semantic system, may permit semantic-visual representations to modulate the \emph{bottom-up} phonological processes as implemented in the current model.

\section{Conclusions}
We conclude that phonological representations mapped dynamically in a \emph{bottom-up} fashion to semantic-visual representations are \emph{sufficient} to capture the early phonological preference effects reported in a visual world task. The semantic-visual preference observed later in such a trial does not require \emph{top-down} feedback from a semantic or visual system. 

We do not claim that such top-down connections do not exist. Indeed, we would expect a proper computational account of the visual world task to include such resources. Our strategy has been to seek to minimise the computational resources needed to account for the phenomenon at hand. We suppose that incremental development of these resources is the best way to achieve understanding of visual world processes.
\bibliographystyle{apacite}

\setlength{\bibleftmargin}{.125in}
\setlength{\bibindent}{-\bibleftmargin}

\bibliography{mduta_cogsci20}

\end{document}